\documentclass[journal]{IEEEtran}

\ifCLASSINFOpdf
\usepackage{amsmath}
\else

\fi

\usepackage{xcolor}
\usepackage{graphicx}
\usepackage{cite}

\usepackage{amsmath}
\usepackage{amsmath}
\usepackage{amssymb}
\usepackage{url}
\usepackage{hyperref}
\usepackage[utf8]{inputenc}
\usepackage[T1]{fontenc}

\usepackage{fancyhdr}  

\fancypagestyle{plain}{
    \fancyhf{}  
    \fancyfoot[C]{\footnotesize This paper has been accepted for publication at the IEEE International Conference on Image Processing (ICIP) 2024 \\
    \thepage}  
}

\begin{document}

\title{MODIPHY: Multimodal Obscured Detection for IoT using PHantom Convolution-Enabled Faster YOLO}

\author{
  \IEEEauthorblockN{
    Shubhabrata Mukherjee\IEEEauthorrefmark{1},
    Cory Beard\IEEEauthorrefmark{0},
    Zhu Li\IEEEauthorrefmark{0}
  }\\
  \IEEEauthorblockA{
    \textit{School of Science and Engineering},\\
    \textit{University of Missouri Kansas City},\\
    Kansas City, USA\\
    \href{mailto:smpw5@umsystem.edu}{smpw5@umsystem.edu};\quad
    \href{mailto:beardc@umkc.edu}{beardc@umkc.edu};\quad
    \href{mailto:zhu.li@ieee.org}{zhu.li@ieee.org}
  }
}

\maketitle

\vspace{2em}

\thispagestyle{plain}

\begin{abstract}

Low-light conditions and occluded scenarios impede object detection in real-world Internet of Things (IoT) applications like autonomous vehicles and security systems. While advanced machine learning models strive for accuracy, their computational demands clash with the limitations of resource-constrained devices, hampering real-time performance. In our current research, we tackle this challenge, by introducing ``YOLO Phantom", one of the smallest YOLO models ever conceived. YOLO Phantom utilizes the novel Phantom Convolution block, achieving comparable accuracy to the latest YOLOv8n model while simultaneously reducing both parameters and model size by 43\%, resulting in a significant 19\% reduction in Giga Floating-Point Operations (GFLOPs). YOLO Phantom leverages transfer learning on our multimodal RGB-infrared dataset to address low-light and occlusion issues, equipping it with robust vision under adverse conditions. Its real-world efficacy is demonstrated on an IoT platform with advanced low-light and RGB cameras, seamlessly connecting to an AWS-based notification endpoint for efficient real-time object detection. Benchmarks reveal a substantial boost of 17\% and 14\% in frames per second (FPS)
for thermal and RGB detection, respectively, compared to the baseline YOLOv8n model. For community contribution, both the code and the multimodal dataset are available on GitHub.\footnote{The code and dataset are available at \url{http://tinyurl.com/46y5twku}}

\end{abstract}

\begin{IEEEkeywords}
Low light object detection, Multimodal fusion, IoT, YOLO, Phantom Convolution
\end{IEEEkeywords}

\IEEEpeerreviewmaketitle

\section{Introduction}
Robust object detection under low light and occlusion is crucial for various applications, including traffic monitoring, public safety, and environmental monitoring. Instances such as Tesla's autopilot oversight and GM's Cruise pedestrian crash underscore safety concerns in autonomous vehicles~\cite{ap_tesla_crash_2023,reuters_gm_recall_2023}. The challenges extend to facial recognition systems, facing accuracy drops in inadequate lighting, necessitating advancements in object detection technologies~\cite{al2022comparing,ahmed2021survey}. AI plays a pivotal role in addressing these low-light challenges, employing innovative approaches like integrating infrared sensors to perceive thermal signatures in darkness. The multimodal augmentation~\cite{josi2023multimodal} of thermal and RGB datasets in multimodal imaging emerge as a powerful solution, particularly in overcoming occlusion. Fusing thermal, visible light, LiDAR, and radar distance scanning data through machine learning significantly improves the performance of object detection and classification systems.

Recent strides in low-light object detection, utilizing deep learning and fusion techniques~\cite{chen2021exploring,kvyetnyy2017object,ye2023llod,wang2022low}, exhibit promise. YOLO-based methods, in particular, have gained acclaim~\cite{liu2022image,yin2023pe,wang2023improved}. While deploying neural network models on edge devices like UAVs presents challenges~\cite{nikouei2018real}, our work bridges this gap by enhancing the YOLOv8 nano~\cite{Jocher_Ultralytics_YOLO_2023}  architecture, giving rise to the ``YOLO Phantom." Tailored for resource-constrained IoT devices, this adaptation ensures compatibility with small-scale hardware, achieving substantial speed improvements for real-time object detection in low-light conditions. Our presentation delves into these enhancements, scrutinizes their performance gains, and showcases their effectiveness in pertinent low-light object detection tasks.

In the context of our current research, we have developed a compact yet efficient multimodal object detection model optimized for effective low-light object detection. Leveraging the FLIRV2 dataset~\cite{flir_v2_dataset}, we crafted a multimodal dataset to train our own YOLO Phantom model, complemented by integrating a NoIR camera Module~\cite{noir} with Raspberry Pi for exceptional performance in low-light scenarios. The distinctive contributions of this paper include:

\begin{itemize}
    \item \textbf{Creation of a compact YOLO model}: Designing one of the smallest YOLO models, by using only 50\% of the kernels in three of the layers, demonstrating superior speed without compromising accuracy.
    
    \item \textbf{YOLO Phantom for multimodal detection}: Introducing YOLO Phantom, capable of detecting multiple modalities such as RGB and thermal. The model underwent training using various modalities and demonstrates effectiveness in scenarios characterized by low light conditions and occlusion.
    
    \item \textbf{Innovative convolution block:} Proposing the novel ``Phantom Convolution" block, by using Depthwise Separable Convolution~\cite{chollet2017xception} and Group Convolution~\cite{krizhevsky2012imagenet} layers we created a faster and more efficient version of Ghost Convolution~\cite{han2020ghostnet}.
    
    \item \textbf{Realization of an end-to-end notification system}: Realizing the new model through a practical end-to-end object detection framework. This involves integrating a Raspberry Pi with RGB and NoIR cameras for simultaneous detection, along with an AWS cloud-based real-time notification system.
\end{itemize}

\section{Implementation strategies}

Customized detection systems power diverse applications, from self-driving cars to surveillance, tailoring to specific environments and tasks. Key types include:
\begin{itemize}
    \item \textbf{Aerial Detection System (ADS):} ADS is utilized in air defense, public safety, disaster recovery~\cite{10025436} and vegetation monitoring, with ground-based installations strategically placed and air-based systems integrated into manned or unmanned aerial vehicles (UAVs) for enhanced situational awareness.

    \item \textbf{Marine Detection System (MDS):} MDS is designed to detect and identify threats to maritime security, using radar, sonar, acoustic sensors, and smart cameras to track vessels and activities in surrounding waters~\cite{zhang2021lightweight}. AI-based software and ML algorithms analyze sensor data for threat identification, with examples like an Automatic Identification System (AIS) for vessel tracking and Long-Range Identification and Tracking (LRIT) for long-distance identification.

    \item \textbf{Ground-based Detection System (GDS):} GDS focuses on detecting, classifying, or tracking objects in terrestrial environments, employing technologies like ground-based radar, RGB or infrared-sensor-based cameras, and motion sensors to identify people, vehicles, or object movement~\cite{9506829}. AI-based infrastructure analyzes collected data for threat identification, including perimeter intrusion detection systems and object detection with facial recognition using smart cameras and machine learning.
\end{itemize}

\begin{figure}[hbt!]
\centerline{\includegraphics[width=0.95\columnwidth]{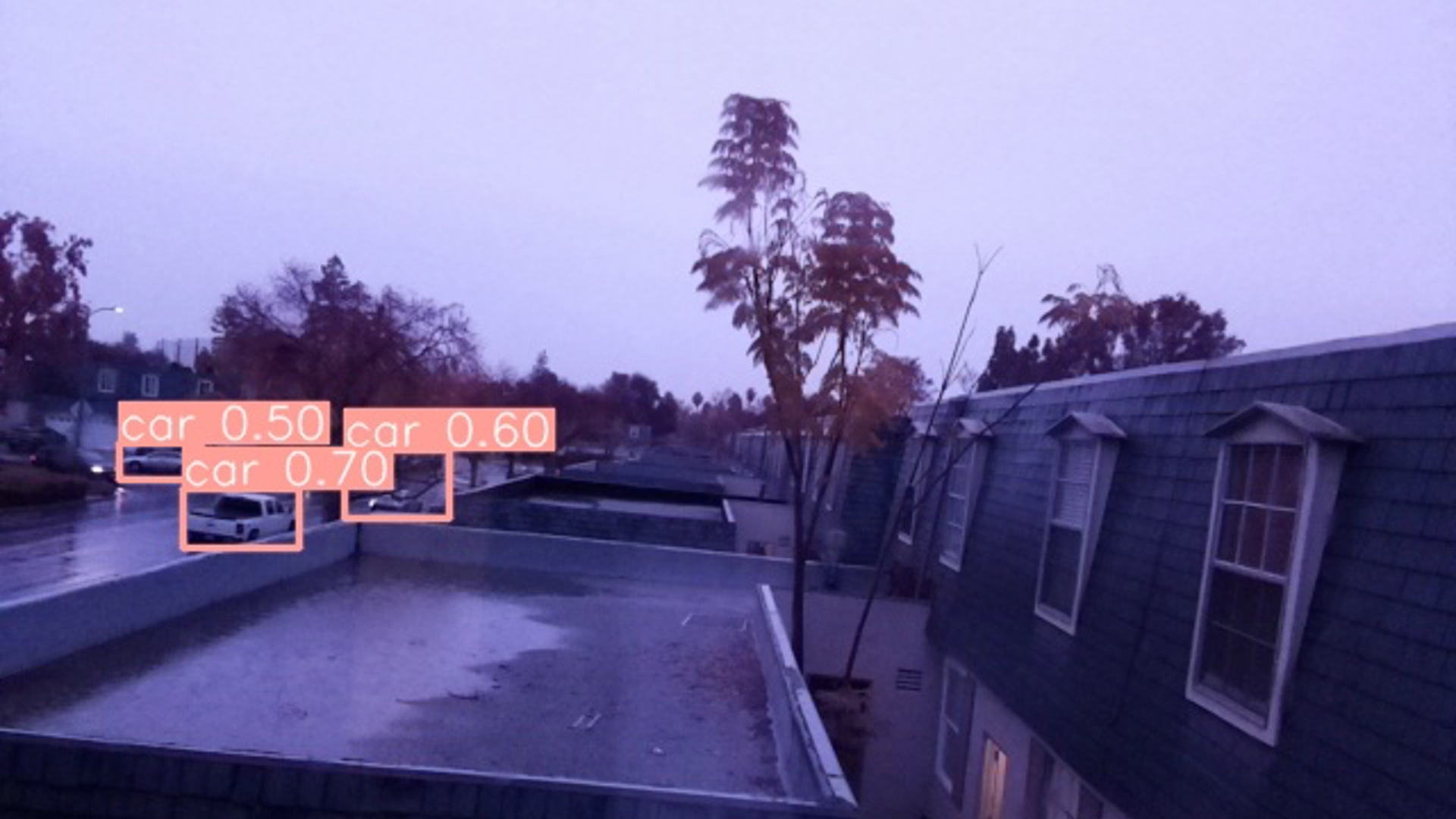}}
\caption{Detection on a Rainy, Obscured Evening with Severe Occlusion Using a Multimodal YOLO Model and a NoIR Camera on a Raspberry Pi Platform}
\label{dark}
\end{figure}

\section{Role of resource optimization}

\begin{figure}[hbt!]
\centerline{\includegraphics[width=0.5\textwidth]{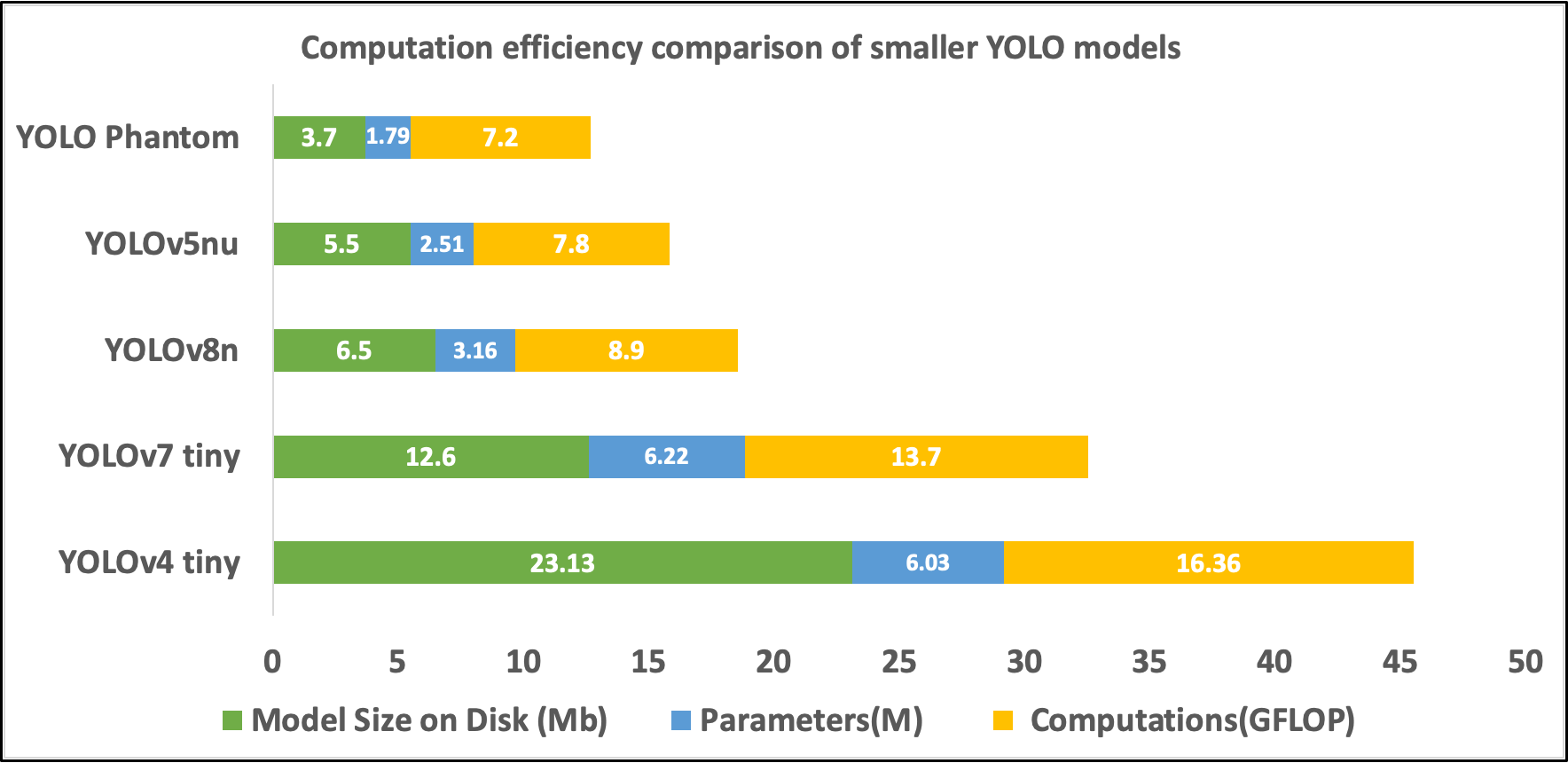}}
\caption{Size, Parameters, and GFLOP comparison of smaller YOLO models}
\label{gflop}
\end{figure}

Limited resources in remote areas often constrain resilient object detection in IoT systems. Optimized resource usage becomes vital for their extended, reliable operation in such environments, enhancing lifespan, accuracy, and overall efficacy. In particular, resource optimization is focused on the efficient utilization of computation resources such as power, memory, and temperature control. Limited power in devices like Unmanned Aerial Vehicle (UAV) or surveillance cars hinders training and running computation-intensive AI models, compromising efficiency, longevity, and overall performance. Through significant improvements, particularly the incorporation of our newly proposed phantom convolution block, we successfully developed one of the most compact YOLO models to date. As illustrated in Fig.~\ref{gflop}, our model boasts the fewest parameters, minimal GFLOPs, and the smallest size when compared to some of the most recent compact YOLO models. Performance evaluation revealed comparable or superior performance in different scenarios compared to other state-of-the-art (SOTA) smaller models, encompassing both RGB and thermal modalities. YOLO Phantom's lightweight design makes it highly suitable for real-time object detection on edge devices.

\section{Our approach: YOLO Phantom}

We have modified the architecture of the smallest version of the Ultralytics YOLOv8 model, YOLOv8n, to enable it to detect both RGB and thermal modality images. Furthermore, we strategically restructured and resized select network blocks to facilitate expedited model inference. And, finally, we trained it using transfer learning with a multimodal RGB and thermal image dataset as described later. This enabled YOLO Phantom to detect multimodal objects much faster than the Ultralytics YOLOv8, with similar or slightly higher accuracy (mAP), as seen in Section ~\ref{Sec:Results}.

\subsection{Ultralytics YOLOv8 architecture}

\begin{figure}[hbt!]
\centerline{\includegraphics[width=0.5\textwidth]{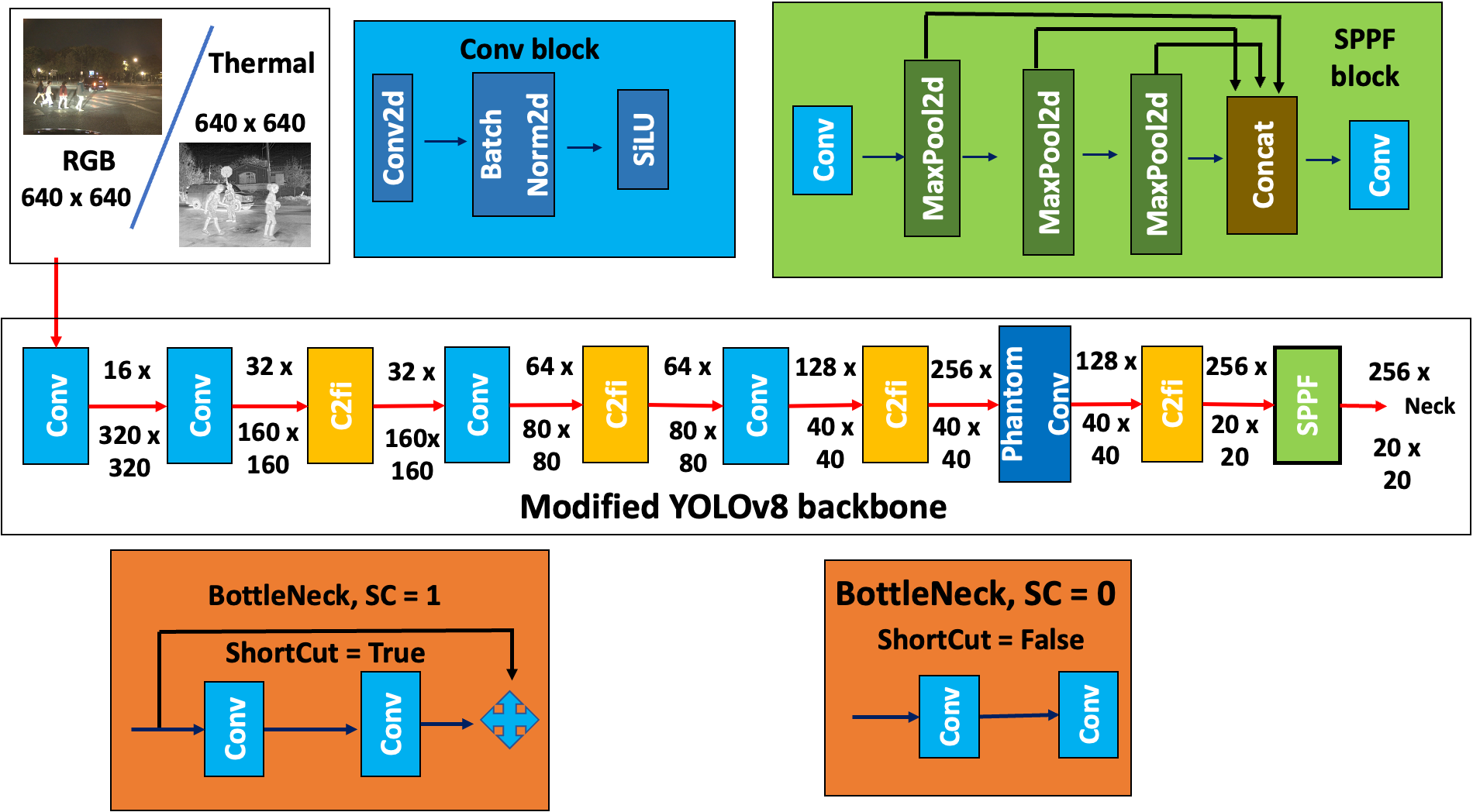}}
\caption{Modified YOLOv8 Backbone~\cite{RangeKingGitHub}}
\label{bbn}
\end{figure}

The Ultralytics YOLOv8 architecture is primarily divided into three main parts: the Backbone, Neck, and Head. The Backbone is responsible for generating feature maps and incorporates the following components: C2f, a modified version of Darknet53 with Cross-Stage Partial (CSP) connections for improved feature re-usability and gradient flow, convolutional layers with varying kernel sizes. The output of the Backbone is feature maps at multiple scales, capturing both low-level details and high-level semantic information. Fig.~\ref{bbn} shows the implementation and our changes in YOLOv8 backbone, which includes the C2f replaced by improved C2f (``C2fi") and new Phantom Convolution blocks. Ultralytics YOLOv8 backbone also consists of SPPF (Spatial Pyramid Pooling - Fast) which enhances the receptive field and captures multi-scale features through parallel pooling branches with different kernel sizes and concatenation of pooled features.

Fig.~\ref{head} shows the Ultralytics YOLOv8 implementation of the neck and head, along with our introduced Phantom Convolution block. Upsample blocks are used to increase the feature resolution and match with C2fi (C2f with ShortCut False) blocks using Concat blocks. The outputs from C2fi goes to Convolution blocks as well as detection heads, used for detecting small, medium, or large objects. It uses a decoupled head to perform various tasks like detection, classification, segmentation separately. YOLOv8 also uses SiLU (Swish Linear Unit) as default activation function, Mosaic augmentation, anchor-free detection, batch normalization for enriched training, and flexible prediction. Please note that inside the Ultralytics repository, in the main architecture file for YOLOV8, the head and neck sections have been collectively referred to as `head'.

\subsection{Architecture improvement for faster and better inference}

In the development of YOLO Phantom, we undertook modifications to accommodate both RGB and thermal image detection, enhancing the model's versatility. Ultralytics YOLOv8 training was conducted on the COCO dataset~\cite{cocodataset}, a predominantly RGB dataset. It was necessary for us to make adjustments to architectural elements and parameter scaling by directing our efforts toward achieving comparable mean average precision (mAP) performance with reduced model complexity and improved speed. Various novel convolution methods have been adapted by researchers to improve accuracy performance (better mAP) or make the model faster (higher FPS) to perform different real-time computer vision tasks. We experimented with quite a few of them for our training; we have mentioned our observations from using them in our model architecture.

\paragraph{\textbf{Group Convolution}}
 Group convolution~\cite{su2020dynamic} divides filters into groups, each operating on a subset of input channels. Unlike traditional convolution connecting to all input channels, group convolution segregates filters, reducing computational costs and enabling parallelization. It enhances efficiency without compromising representational power, especially in models with many channels. The group convolution operation can be expressed as below:

 \begin{figure}[hbt!]
 \centerline{\includegraphics[width=0.5\textwidth]{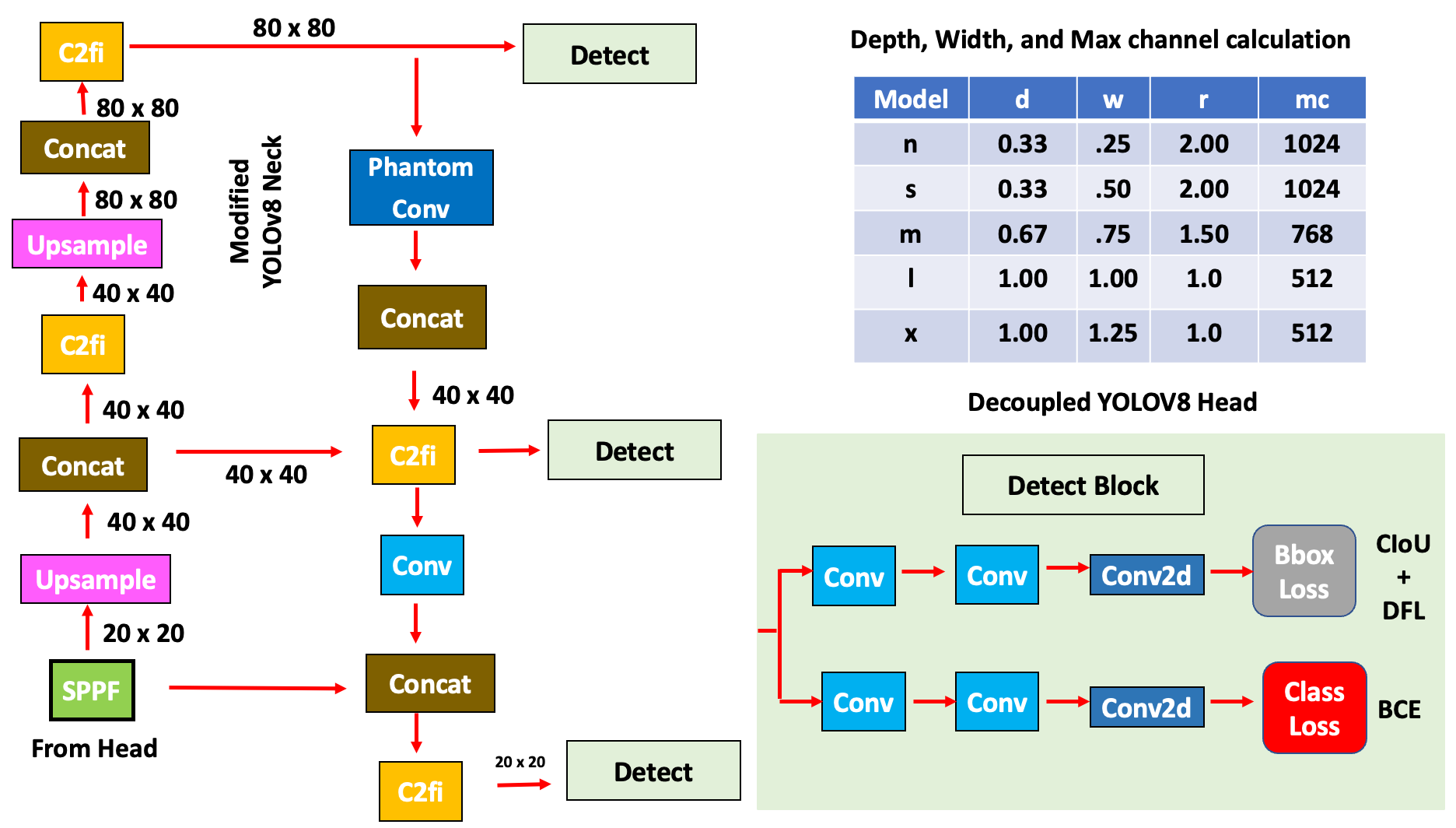}}
 \caption{Modified YOLOv8 Neck and Decoupled detection Head~\cite{RangeKingGitHub}}
 \label{head}
 \end{figure}

\begin{itemize}
\item Input tensor: $X \in \mathbb{R}^{N \times C \times H \times W}$
\item Weight tensor: $W \in \mathbb{R}^{G \times \frac{C}{G} \times K \times K}$
\item Bias term: $b \in \mathbb{R}^{G \times 1}$
\item Output tensor: $Y \in \mathbb{R}^{N \times G \times \frac{C}{G} \times H' \times W'}$
\end{itemize}

\begin{equation}
Y_{n, g, i, j} = \sum_{c=1}^{\frac{C}{G}} X_{n, (g-1)\frac{C}{G} + c, i', j'} \cdot W_{g, c, k', l'} + b_g
\end{equation}

\paragraph{\textbf{Depth-wise separable Convolution}}

Depth-wise Separable convolution~\cite{chollet2017xception} is a CNN variant that splits a standard convolution into two steps: depth-wise convolution, where each input channel is convolved independently, and point-wise convolution, combining results across channels with 1x1 filters. This separation significantly reduces computational complexity and parameters compared to traditional convolutions, making it widely adopted in mobile and edge devices for efficiency without compromising expressiveness. It strikes a balance between computational efficiency and model performance, making it suitable for resource-constrained environments.

\begin{itemize}
\item Input tensor: $X \in \mathbb{R}^{N \times C \times H \times W}$
\item Depthwise convolution filter: $D \in \mathbb{R}^{C \times 1 \times K_d \times K_d}$
\item Pointwise convolution filter: $P \in \mathbb{R}^{C' \times C \times 1 \times 1}$
\item Bias term: $b \in \mathbb{R}^{C' \times 1}$
\item Output tensor: $Y \in \mathbb{R}^{N \times C' \times H' \times W'}$
\end{itemize}

Depthwise Convolution:

\begin{equation}
Y_{dw} = \text{DepthwiseConv}(X, D)  // Y_{dw} \in \mathbb{R}^{N \times C \times H' \times W'}
\end{equation}

Pointwise Convolution:

\begin{equation}
Y = \text{PointwiseConv}(Y_{dw}, P) + b  // Y \in \mathbb{R}^{N \times C' \times H' \times W'}
\end{equation}
Now, if we combine above two operations:

\begin{align}
Y_{n, c', i, j} = &\sum_{c=1}^{C} \sum_{k'=1}^{K_d} \sum_{l'=1}^{K_d} X_{n, c, i \times S + k', j \times S + l'} \cdot D_{c, 1, k', l'} \nonumber \\
& + \sum_{c=1}^{C} Y_{dw, n, c, i, j} \cdot P_{c', c, 1, 1} + b_{c'}
\end{align}

\paragraph{\textbf{Ghost Convolution}}
Ghost Convolution~\cite{han2020ghostnet} enhances convolutional neural networks (CNNs) by introducing a parallel ``ghost branch" alongside the main convolutional layers. The ghost branch, created by subsampling input channels, processes a reduced input version and combines its output with the main branch. This technique balances computational efficiency and model expressiveness. By utilizing fewer channels in the ghost branch, Ghost Convolution achieves resource savings while retaining essential features. It is particularly valuable in scenarios with limited computational resources like mobile devices where efficiency is crucial.

\begin{figure}[hbt!]
\centerline{\includegraphics[width=0.95\columnwidth]{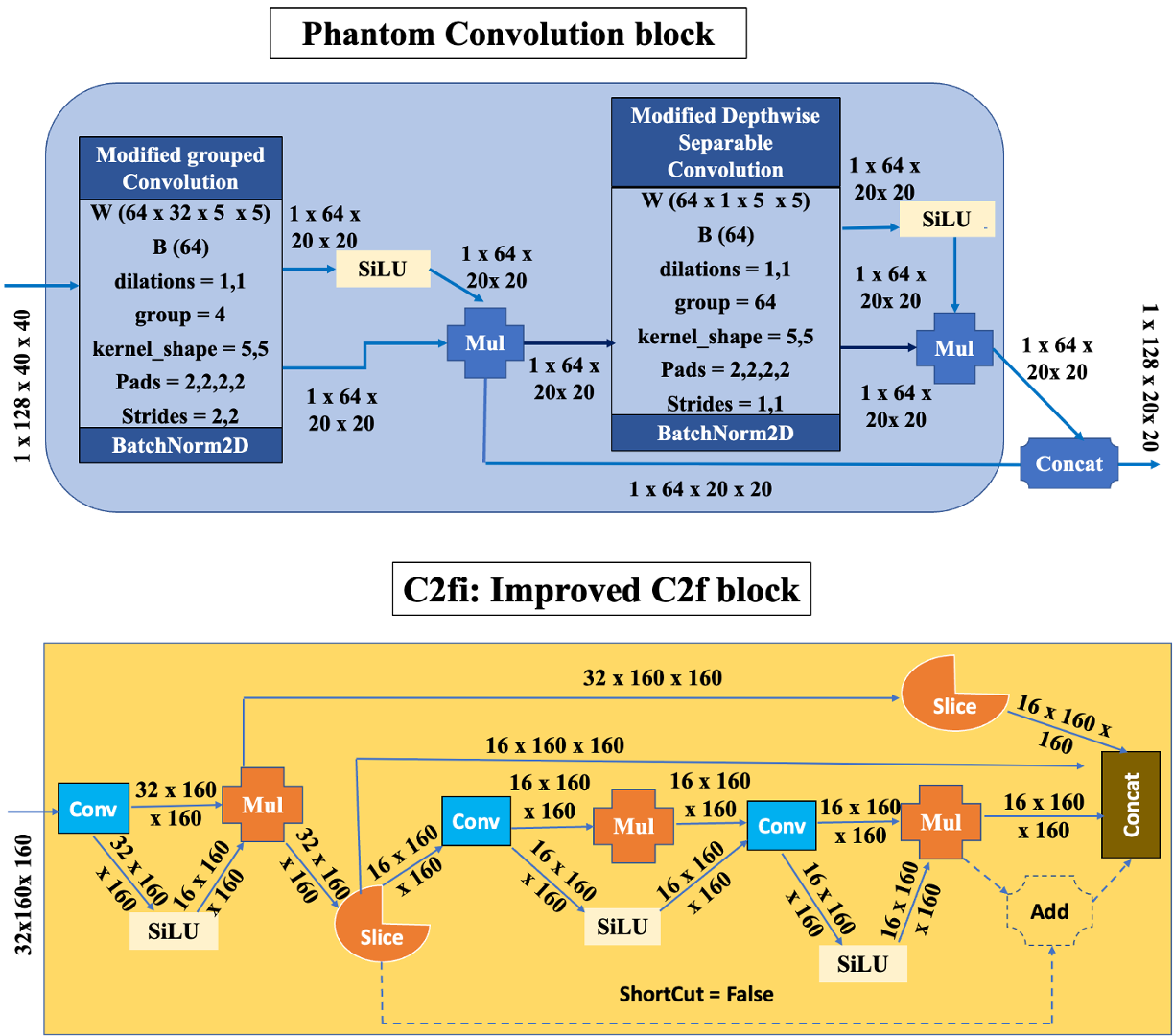}}
\caption{Phantom Convolution and C2fi Block Architecture}
\label{newblocks}
\end{figure}

Various strategies, culminating in adopting Ghost Convolution, were employed to enhance feature extraction with fewer parameters. As seen in Fig.~\ref{newblocks}, the Ghost Convolution block underwent re-architecting, with the first layer transformed into a group convolution (group size 4) for increased operational speed and larger kernel sizes (5x5) to expand the receptive field. The second layer utilized Depth-wise Separable Convolution for efficiency, leading to a compact block with improved computational efficiency, named ``Phantom Convolution".

The C2f blocks were modified by eliminating forward pass connections, introducing ``C2fi" or ``C2f improved" blocks for a smaller, faster, and more efficient model. These blocks, with shortcut connections set to false, replaced traditional C2f blocks throughout the architecture. These changes allowed a reduction in the number of filters in deeper layers, halving the number of filters in layer 7 and in the head of the backbone. Notably, the seventh layer incorporated the Phantom Convolution block, and the final convolution block in the head (layer 19) was replaced with Phantom Convolution to enhance efficiency and expedite the detection process.

\begin{figure}[hbt!]
\centerline{\includegraphics[width=0.8\columnwidth]{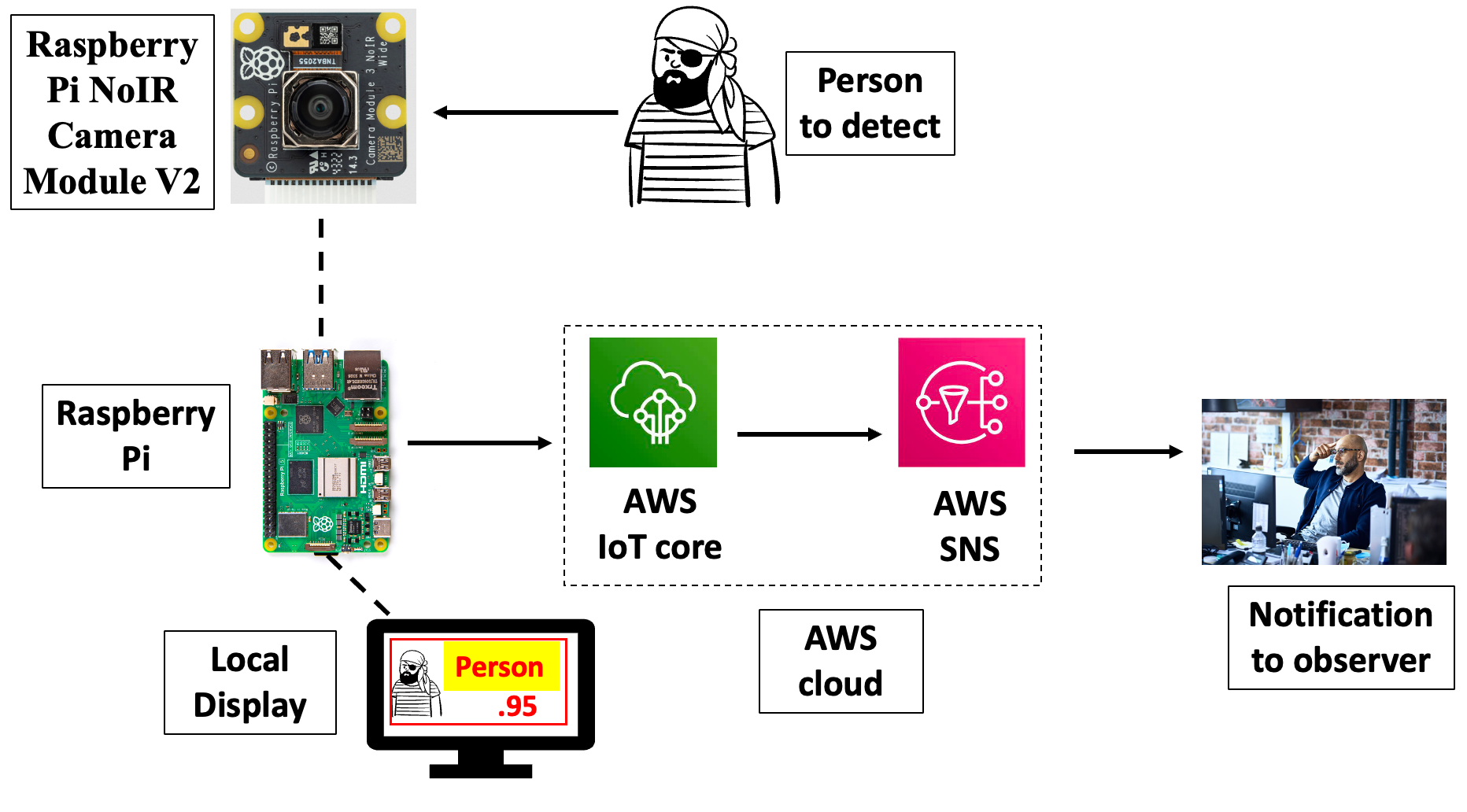}}
\caption{End-to-End Detection and Notification Pipeline }
\label{noti}
\end{figure}

\subsection{Experimental Setup}

\begin{figure}[hbt!]
\centerline{\includegraphics[width=0.4\textwidth]{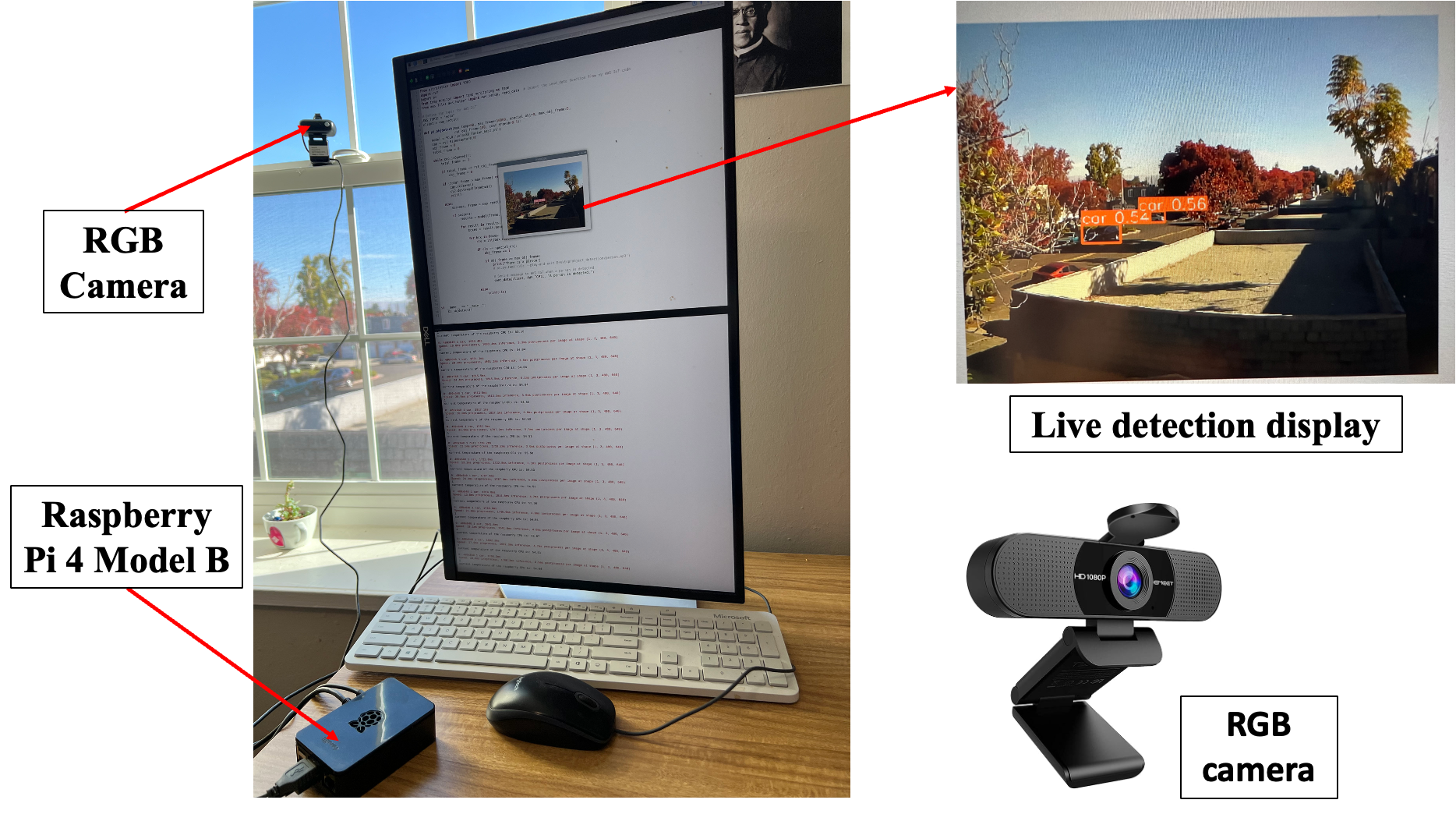}}
\caption{Detection Setup Using Raspberry Pi and RGB Camera}
\label{aws}
\end{figure}

In this study, we devised an end-to-end detection and notification system utilizing IoT-based hardware in conjunction with Amazon Web Services (AWS) for cloud-based notifications (Fig.~\ref{aws}). Specifically, we employed a Raspberry Pi~\cite{raspberrypi} version~4, model~B (CanaKit Extreme, 128 GB, 8GB, BullsEye OS), as the hosting and execution platform for the object detection model. An external RGB camera was interfaced with the USB port of the Raspberry Pi to capture relevant data. We also connected a NoIR Camera Module V2 (Fig.~\ref{noti}) to the Raspberry Pi using Camera Serial Interface (CSI). This camera module can see in low-light without an infrared filter, making it ideal for low-light detection. To establish connectivity between the Raspberry~Pi and the cloud, we leveraged an AWS IoT Core, thereby constructing an event notification framework. This integration empowered us to dispatch customized notifications based on detected events; in our experimentation, we validated this functionality by triggering notifications upon the detection of a person. Furthermore, we used AWS Simple Notification Service (SNS), to augment the AWS IoT Core in disseminating alerts. This feature facilitated the transmission of notifications to designated email addresses, phone numbers, or mobile applications, enhancing the versatility and accessibility of our alerting mechanism.

\begin{figure*}[hbt!]
\centerline{\includegraphics[width=0.95\textwidth]{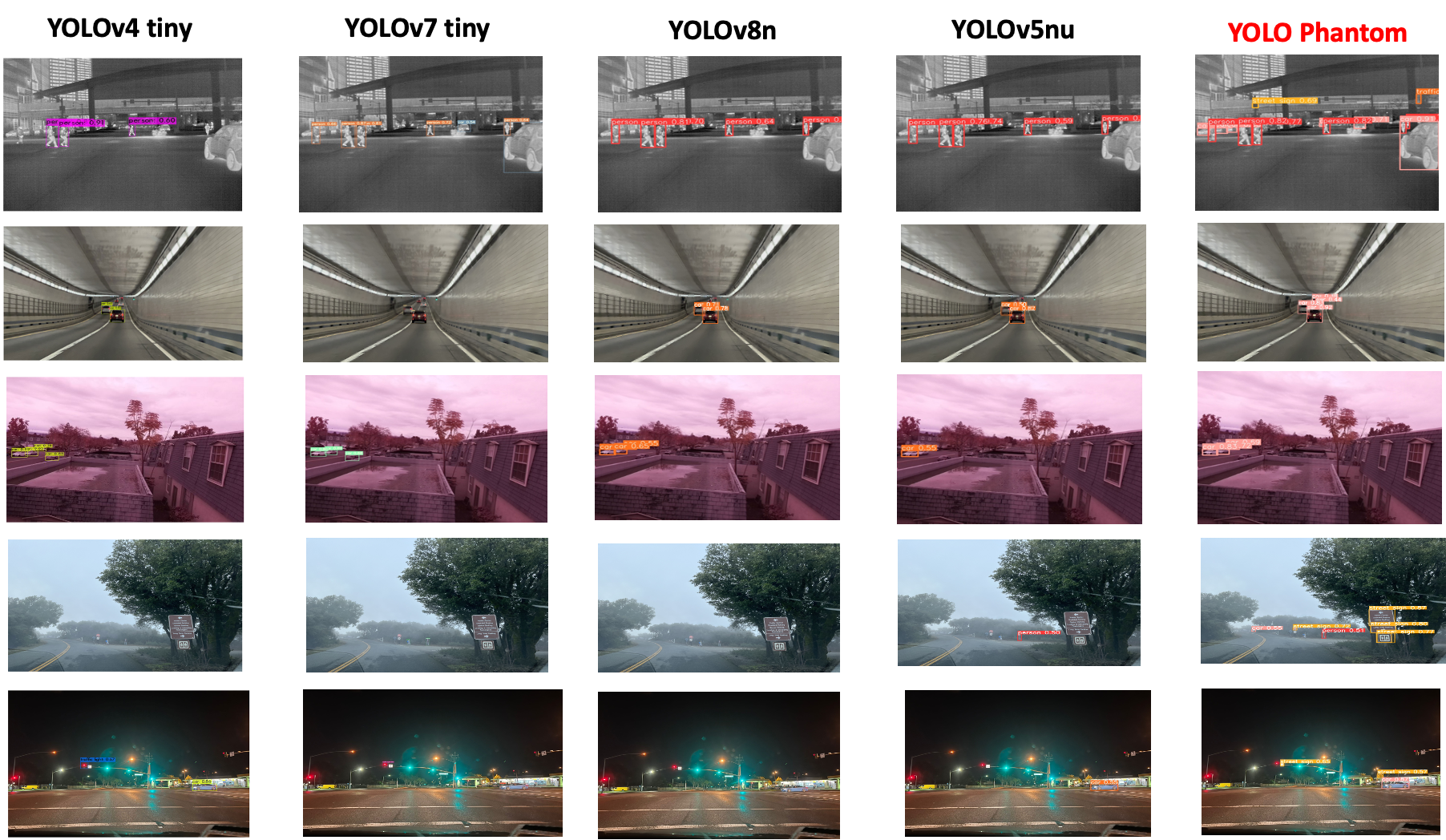}}
\caption{Various Occluded and Low-Light Detection Scenario Comparison with Smaller YOLO Models (left to right) Using NoIR or RGB Cameras}
\label{ous}
\end{figure*}

To bolster the core detection framework, we integrated supplementary safety controls. A temperature monitoring system for the Raspberry Pi automatically halts detection when exceeding a conservative 60°C threshold. Additionally, an audio alert system triggers upon person detection, offering redundancy beyond visual notifications. These enhancements increase responsiveness and provide layered alerts for enhanced situational awareness and threat mitigation.

\subsection{Dataset description and Training}
Utilizing the Teledyne FLIR ADAS Thermal Dataset v2~\cite{flir_v2_dataset}, we curated a multimodal dataset comprising both RGB and thermal imagery. The capability to perceive thermal infrared radiation offers both complementary and distinct advantages compared to conventional sensor technologies like visible cameras, Lidar, and radar systems. Teledyne FLIR thermal sensors excel in detecting and classifying objects under challenging conditions, including total darkness, fog, smoke, adverse weather, and glare. The primary dataset comprises 26,442 annotated images spanning 15 categories. From the training and validation sets, we selected 10,478 thermal and 10,053 RGB images for the training set, and a combination of 1,128 thermal and 1,069 RGB images for validation to construct the multimodal dataset for training. Focusing on the four most represented classes (person, car, traffic light, street sign), we conducted experiments. Additionally, a separate thermal test dataset of 3,749 images and an RGB test dataset of 3,493 images were constructed from the main FLIR~V2 dataset's test section for benchmarking purposes. Notably, in the original FLIR V2 dataset, the RGB and thermal images are not directly correspondent; there is no thermal image for the exact scene depicted in an RGB image or vice-versa. Consequently, our curated multimodal dataset mixes various thermal and RGB images captured in different scenes. This helps the model generalize better and avoids misalignment between RGB and thermal images of the same scene.

YOLO Phantom was trained using our multimodal dataset on a CentOS cluster with two ``NVIDIA RTX A6000" GPUs and 64 CPUs for 100 epochs. Images were resized to 640x640 pixels. The ``yolov8n.pt" pre-trained weights served as the starting point for the training process.

\section{Results and Analysis}
\label{Sec:Results}
\subsection{Out of sample testing}
This section presents a performance comparison of YOLO Phantom against other state-of-the-art (SOTA) smaller YOLO models. The evaluation employed a diverse image set encompassing low-light and occlusion scenarios (thermal imaging, tunnel interiors, cloudy evenings, foggy mornings, nighttime). Both RGB and NoIR cameras captured the images. All the detection has been performed with a non-max-suppression (NMS) 0.5 threshold. Figure~\ref{ous} showcases YOLO Phantom's competitive performance, matching or exceeding smaller models like YOLOv4-tiny, YOLOv7-tiny, YOLOv8n, and YOLOv5nu, despite having 4-2x fewer parameters.

\subsection{Performance on different modality data}
In this section, we conducted a comprehensive benchmark analysis of YOLO Phantom against recent small YOLO models, including YOLOv5nu, Ultralytics YOLOv8n, and YOLOv8n trained on multimodal data without any architecture modifications. The comparison was performed on two separate modalities: RGB and thermal data. We considered two key performance metrics for benchmarking: accuracy (mAP50-95(B)) and frames per second (FPS). The mean Average Precision (mAP) values were calculated using the COCO evaluation metrics~\cite{lin2014microsoft}, which measure the area under the precision-recall curve for each class, averaging these values, and then averaging across all classes. The mAP scores reflect the model's accuracy in detecting objects at various Intersections over Union (IoU) thresholds. We used the ``benchmark" function from the Ultralytics YOLO repository to calculate the benchmarking scores across five different platforms: PyTorch, TorchScript, ONNX, OpenVINO, and ncnn, and the benchmarking process was conducted on an M1 MAC OS CPU.

\begin{figure}[hbt!]
\centerline{\includegraphics[width=0.5\textwidth]{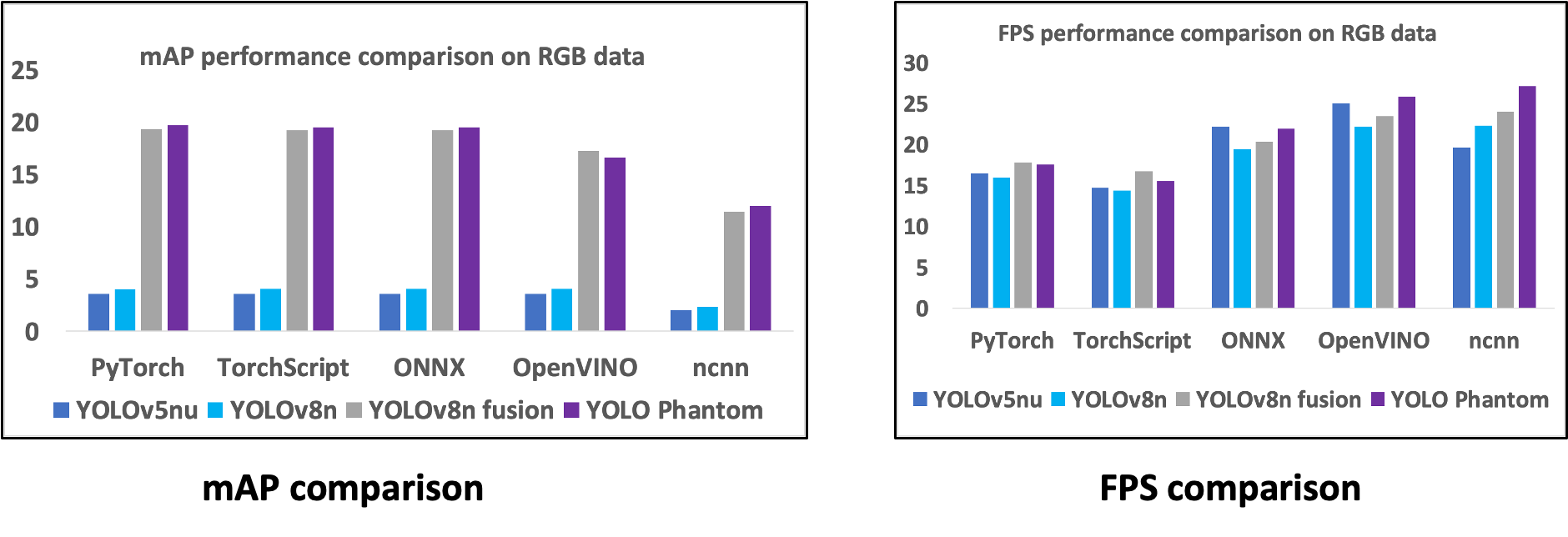}}
\caption{mAP and FPS performance comparison on RGB data with pretrained YOLOV5nu, pretrained YOLOv8n, YOLOv8n transfer learned on multimodal data, and YOLO Phantom}
\label{rgb}
\end{figure}

\begin{figure}[hbt!]
\centerline{\includegraphics[width=0.5\textwidth]{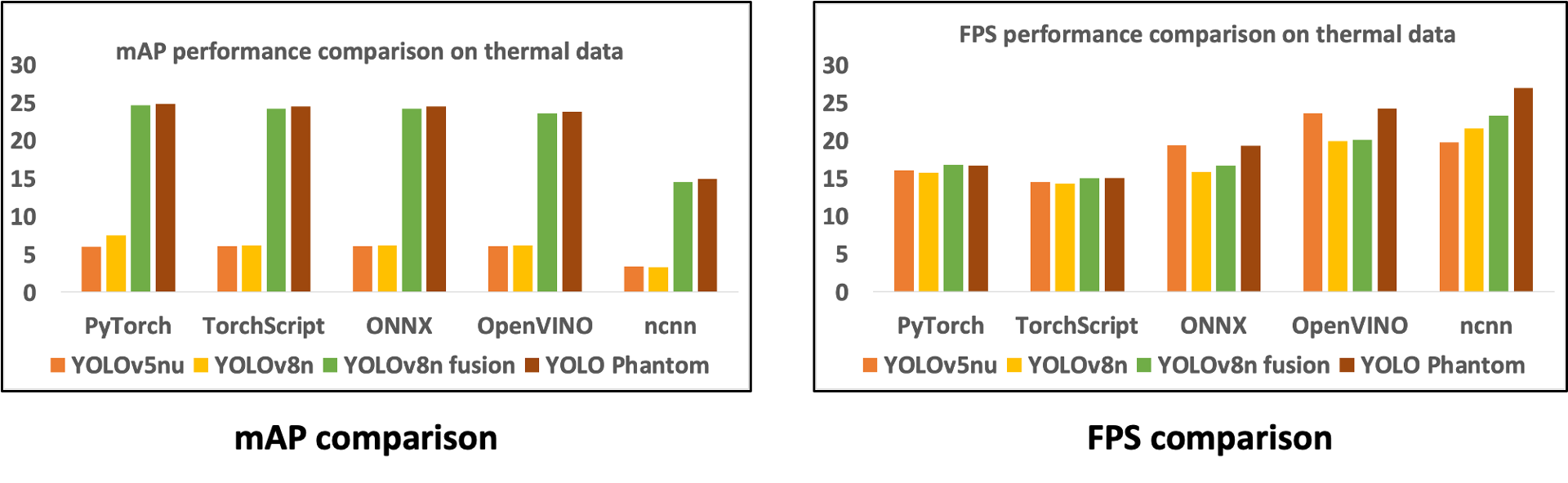}}
\caption{mAP and FPS performance comparison on thermal data with pretrained YOLOV5nu, pretrained YOLOv8n, YOLOv8n transfer learned on multimodal data, and YOLO Phantom}
\label{thermal}
\end{figure}

As depicted in Fig.~\ref{rgb}, YOLO Phantom demonstrates a significantly superior performance compared to pre-trained models such as YOLOv5nu and YOLOv8n. It exhibits a slightly higher accuracy than the YOLOv8n fusion model trained on our multimodal data. Moreover, YOLO Phantom consistently achieves higher FPS across various model variants, surpassing both pre-trained smaller models and the YOLOv8n fusion model. Similar trends were observed in Fig.~\ref{thermal} during the thermal data benchmarking. YOLO Phantom outperforms its competitors in mAP and achieves superior speed due to its smaller size.

\subsection{Cross-modality performance}
This section evaluates the performance of multimodal models against unimodal models using reverse modality data. This approach demonstrably highlights the added value of multimodal learning. Table~\ref{tab1} showcases performance (mAP) on RGB data comparing YOLOv8n trained on thermal, YOLOv8n trained on RGB, and YOLO Phantom trained multimodally. While the thermal-trained YOLOv8n struggled, the RGB-trained one excelled. Notably, YOLO Phantom achieved similar or better performance in most cases, demonstrating its multimodal advantage.

\begin{table}[htbp]
  \caption{Cross-modality mAP performance on RGB data}
  \begin{center}
    \begin{tabular}{|c|c|c|c|}
      \hline
      \cline{2-4}
      \textbf{\textit{Format}} & \textbf{\textit{Thermal Trained}} & \textbf{\textit{RGB Trained}} & \textbf{\textit{YOLO Phantom}} \\
      \hline
      PyTorch & 9.29 & 19.65 & \textbf{19.72} \\
      \hline
      TorchScript & 9.19 & 19.57 & \textbf{19.51} \\
      \hline
      ONNX & 9.19 & 19.57 & \textbf{19.51} \\
      \hline
      OpenVINO & 8.43 & 10.24 & \textbf{16.65} \\
      \hline
      ncnn & 5.22 & 11.99 & \textbf{11.95} \\
      \hline
    \end{tabular}
    \label{tab1}
  \end{center}
\end{table}

In the next case, we performed a similar comparison, but instead of accuracy (mAP) we measured speed (FPS). As observed in Table~\ref{tab2}, in this case, YOLO Phantom performed similar or better than thermal-trained or RGB-trained models.

\begin{table}[htbp]
  \caption{Cross-modality FPS performance on RGB data}
  \begin{center}
    \begin{tabular}{|c|c|c|c|}
      \hline
      \cline{2-4}
      \textbf{\textit{Format}} & \textbf{\textit{Thermal Trained}} & \textbf{\textit{RGB Trained}} & \textbf{\textit{YOLO Phantom}} \\
      \hline
      PyTorch & 17.50 & 17.14 & \textbf{17.53} \\
      \hline
      TorchScript & 16.35 & 15.77 & \textbf{15.50} \\
      \hline
      ONNX & 20.12 & 20.45 & \textbf{21.92} \\
      \hline
      OpenVINO & 22.86 & 23.28 & \textbf{25.86} \\
      \hline
      ncnn & 23.77 & 23.85 & \textbf{27.15} \\
      \hline
    \end{tabular}
    \label{tab2}
  \end{center}
\end{table}

Next, we conducted a cross-modality mAP performance comparison on the thermal dataset. As seen in Table~\ref{tab3}, in this case also the reverse modality model (RGB-trained) performed poorly compared to the same modality (thermal-trained) YOLOv8n model. However, the thermal-trained model performed marginally better compared to YOLO Phantom on this occasion.
\vspace{-0.2 cm}
\begin{table}[htbp]
  \caption{Cross-modality mAP performance on thermal data}
  \begin{center}
    \begin{tabular}{|c|c|c|c|}
      \hline
      \cline{2-4}
      \textbf{\textit{Format}} & \textbf{\textit{RGB Trained}} & \textbf{\textit{Thermal Trained}} & \textbf{\textit{YOLO Phantom}} \\
      \hline
      PyTorch & 7.21 & 25.21 & \textbf{24.82} \\
      \hline
      TorchScript & 6.81 & 24.9 & \textbf{24.46} \\
      \hline
      ONNX & 6.81 & 24.9 & \textbf{24.46} \\
      \hline
      OpenVINO & 2.81 & 24.58 & \textbf{23.79} \\
      \hline
      ncnn & 3.87 & 15.26 & \textbf{14.94} \\
      \hline
    \end{tabular}
    \label{tab3}
  \end{center}
\end{table}
\vspace{-0.2 cm}
Finally, we performed a cross-modality FPS comparison on the thermal dataset. As observed in Table IV, RGB, and thermal modality trained models performed nearly similarly, but YOLO
Phantom shows much better FPS performance compared to both.
\vspace{-0.2 CM}
\begin{table}[htbp]
  \caption{Cross-modality FPS performance on thermal data}
  \begin{center}
    \begin{tabular}{|c|c|c|c|}
      \hline
      \cline{2-4}
      \textbf{\textit{Format}} & \textbf{\textit{RGB Trained}} & \textbf{\textit{Thermal Trained}} & \textbf{\textit{YOLO Phantom}} \\
      \hline
      PyTorch & 17.76 & 16.54 & \textbf{16.65} \\
      \hline
      TorchScript & 15.54 & 15.01 & \textbf{15.09} \\
      \hline
      ONNX & 19.16 & 18.48 & \textbf{19.32} \\
      \hline
      OpenVINO & 22.92 & 22.38 & \textbf{24.22} \\
      \hline
      ncnn & 23.98 & 24.03 & \textbf{26.96} \\
      \hline
    \end{tabular}
    \label{tab4}
  \end{center}
\end{table}
\vspace{-0.5 CM}

\section{Conclusion and future work}
YOLO Phantom opens the door to many new real-world object detection tasks, particularly in resource-constrained environments. This lightweight model exhibits a remarkable equilibrium between rapid execution, elevated accuracy, and resource efficiency – the three most critical factors for IoT devices. The innovative Phantom Convolution block seamlessly integrates into any framework, unlocking superior feature extraction with reduced computational burden. Its novel architecture transcends specific tasks, demonstrating exceptional performance across diverse, resource-scarce detection scenarios. To further strengthen an implementation's practical viability for resource-constrained IoT deployments, user-defined temperature monitoring was added to safeguard against thermal excursions, while optional audio alerts offer additional redundancy for enhanced situational awareness and threat mitigation. This advancement holds promise for broader performance gains across computer vision, encompassing tasks like classification, segmentation, and even generative modeling. The impact of YOLO Phantom extends beyond just enhanced road safety and refined surveillance applications, paving the way for pioneering applications in environmental monitoring, wildlife tracking, and beyond.

\ifCLASSOPTIONcaptionsoff
  \newpage
\fi

\bibliographystyle{IEEEtran}
\bibliography{ref}

\end{document}